\documentclass[twoside]{article} 

\usepackage[accepted]{aistats2024}
\usepackage{amsmath,amsfonts,bm}

\def\eqref#1{equation~\ref{#1}}

\def\1{\bm{1}}

\DeclareMathAlphabet{\mathsfit}{\encodingdefault}{\sfdefault}{m}{sl}
\SetMathAlphabet{\mathsfit}{bold}{\encodingdefault}{\sfdefault}{bx}{n}

\usepackage{hyperref}
\usepackage{url}
\usepackage{graphicx}
\usepackage{multirow}
\usepackage{booktabs}
\usepackage{adjustbox}
\usepackage{siunitx}
\usepackage{amsmath}

\begin{document}

\twocolumn[
\aistatstitle{Unveiling the Potential of Probabilistic \\Embeddings in Self-Supervised Learning}
\aistatsauthor{ Denis Janiak \And Jakub Binkowski \And Piotr Bielak \And Tomasz Kajdanowicz }
\aistatsaddress{ Wrocław University of Science and Technology }
]

\begin{abstract}
In recent years, self-supervised learning has played a pivotal role in advancing machine learning by allowing models to acquire meaningful representations from unlabeled data. An intriguing research avenue involves developing self-supervised models within an information-theoretic framework, but many studies often deviate from the stochasticity assumptions made when deriving their objectives. To gain deeper insights into this issue, we propose to explicitly model the representation with stochastic embeddings and assess their effects on performance, information compression and potential for out-of-distribution detection. From an information-theoretic perspective, we seek to investigate the impact of probabilistic modeling on the information bottleneck, shedding light on a trade-off between compression and preservation of information in both representation and loss space. Emphasizing the importance of distinguishing between these two spaces, we demonstrate how constraining one can affect the other, potentially leading to performance degradation. Moreover, our findings suggest that introducing an additional bottleneck in the loss space can significantly enhance the ability to detect out-of-distribution examples, only leveraging either representation features or the variance of their underlying distribution.
\end{abstract}

\begin{figure}[h]
    \centering
    \includegraphics[width=0.49\textwidth]{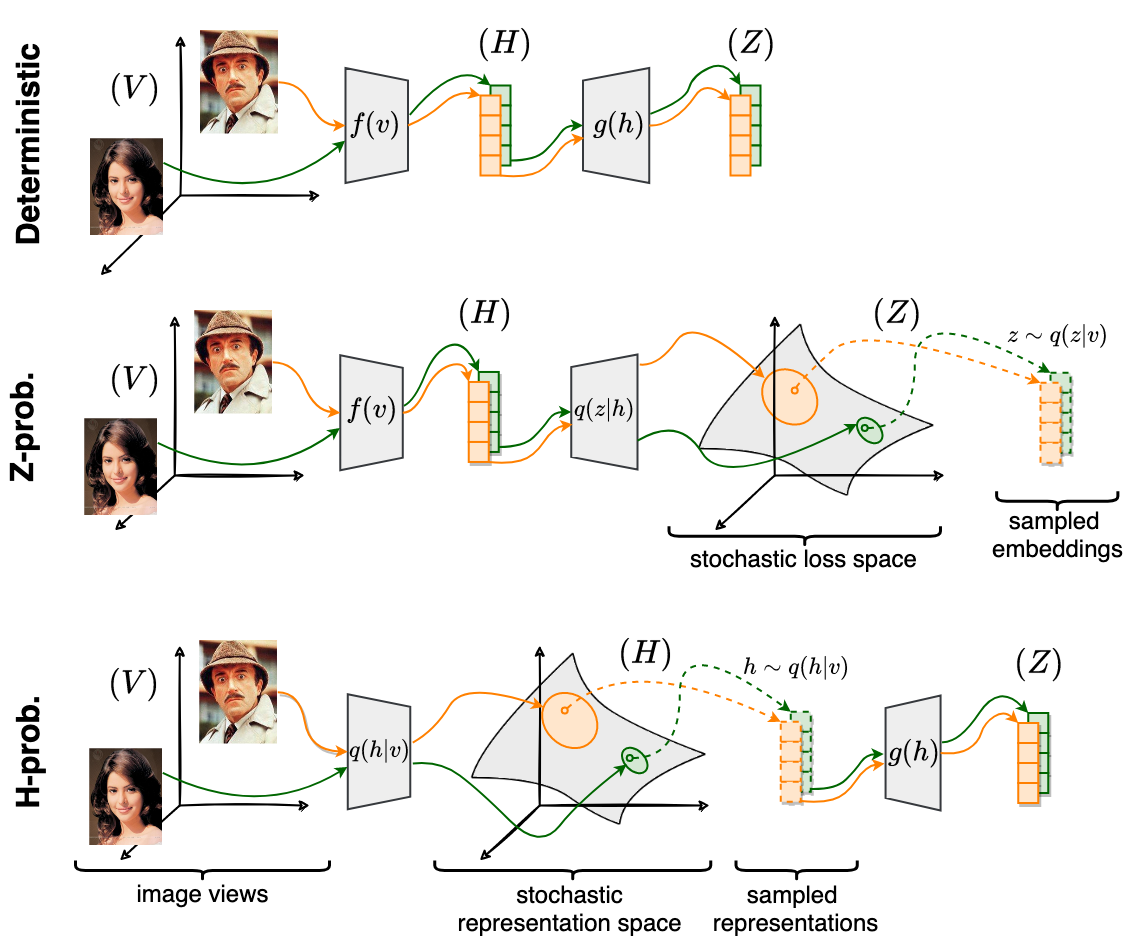}
    \caption{Schematic representation of the multi-view encoding process. Starting with image views (V), the data is transformed through deterministic or stochastic pathways. The deterministic flow processes the input via encoder $f(\cdot)$ into a deterministic representation space (H) before reaching the deterministic loss space (Z). The two probabilistic pathways, (Z-prob.) and (H-prob.), incorporate stochasticity through sampling from associated spaces (representation or loss), generating samples through stochastic encoders $q(h|v)$ and $q(z|v)$. Arrows indicate data flow, with solid lines denoting deterministic pathways and dotted indicating stochastic transformations.}
    \label{fig:bayes_ssl_diagram}
\end{figure}

\section{Introduction}
In recent years, self-supervised learning methods have gained prominence in computer vision, enabling the utilization of abundant unlabeled data \cite{ozbulak_know_2023}.
Contrastive methods \cite{oord_representation_2019, chen_simple_2020, he_momentum_2020}, which train the model to discriminate positive and negative samples from a batch of examples, proved to be very successful in many downstream tasks, facilitating rapid advancement in the domain. However, these methods still have some inherent constraints, such as the necessity of mining negative examples. 

Non-contrastive techniques have addressed this issue through strategies such as feature decorrelation and information maximization \cite{ermolov_whitening_2021, zbontar_barlow_2021, bardes_vicreg_2022}, as well as distillation and architectural constraints \cite{grill_bootstrap_2020, chen_exploring_2021, caron_emerging_2021}.
These approaches have effectively ensured uniformity \cite{wang_understanding_2022} and prevented representation collapse \cite{jing_understanding_2022}.
Methods employing feature decorrelation to prevent representation collapse are of particular interest, as they are closely linked to an information-theoretic framework, which provides a formal means of quantifying the information content and redundancy in data, facilitating a deeper understanding and analysis of these techniques.

In many real-world scenarios, the abundance of data comes with inherent uncertainties. For instance, images captured in the wild might possess inherent aleatoric uncertainties due to factors like low resolution, imperfect cropping, or angle \cite{kirchhof_url_2023}.
Recent machine learning advancements emphasize quantifying these uncertainties, crucial for safety-critical applications, like medical imaging \cite{barbano_uncertainty_2022}, or active learning \cite{settles_active_2009, sharma_evidence-based_2017}.

In self-supervised learning, probabilistic embeddings present a promising avenue to model and leverage such uncertainties effectively,  bridging the gap between data invariance and robust representation. The inherent flexibility of predicting a distribution over the embedding space, instead of deterministic point estimates, enables models to function effectively in environments laden with uncertainty or noise \cite{karpukhin_probabilistic_2022}. 
However, there remain challenges in seamlessly integrating probabilistic embeddings with self-supervised techniques, which often leads studies to deviate from the stochasticity assumptions set out in their objectives \cite{lee_compressive_2021, wang_rethinking_2022, shwartz-ziv_information-theoretic_2023}.
This paper, hence, aims to explore these intricacies, shedding light on the harmonization of probabilistic embeddings with decorrelation-based self-supervised methods and the consequent effects on out-of-distribution detection.

In our paper, we introduce probabilistic embeddings to feature decorrelation-based self-supervised methods (Barlow Twins \cite{zbontar_barlow_2021}, VICReg \cite{bardes_vicreg_2022}) and explore their effect within either the representation or loss space. 
Our contributions are as follows:
\begin{itemize}
    \item We demonstrate that using probabilistic embeddings in the loss space (Z) produces results equivalent to deterministic methods; while using them in the representation space (H) results in a bottleneck, adversely affecting downstream task performance.
    \item We examine the mutual information between input, representation, and loss spaces and conjecture that the aforementioned decline is caused by the representation bottleneck that prioritizes data invariances while compromising generalization. Empirical evaluations support our hypothesis.
    \item We showcase our method's capability in detecting out-of-distribution samples using the variance of the embedding distribution, outperforming both label-free (Mahalanobis) and label-based (MaxSoftmax, ODIN) detectors.
\end{itemize}

\section{Related works}
\label{sec:related_works}
\paragraph{Self-supervised learning (SSL)} The primary objective of SSL is to optimize a specific loss function tailored to capture meaningful patterns or relationships within unlabeled data. This loss function is crafted to create surrogate tasks, such as predicting missing parts of the data, rotations, or other data transformations, that encourage the network to learn useful and invariant features from the input data. 
For instance, in contrastive learning (\cite{chen_simple_2020}, \cite{oord_representation_2019}), the goal is to maximize the agreement between positive (similar) pairs of data samples while minimizing it for negative (dissimilar) pairs. This strategy prompts the network to ensure that similar data instances have representations close to each other in a high-dimensional space, simultaneously pushing dissimilar samples apart.
On the other hand, non-contrastive methods adopt various forms of mechanism to prevent representation collapse, eliminating the need for negative samples \cite{shwartz-ziv_compress_2023}. This could be some architectural constraints \cite{grill_bootstrap_2020, chen_exploring_2020}, clustering-based objective \cite{caron_unsupervised_2021}, or feature decorrelation \cite{zbontar_barlow_2021, bardes_vicreg_2022}. 
In our investigation, we concentrate on the feature decorrelation-based methods, i.e., Barlow Twins \cite{zbontar_barlow_2021} and VICReg \cite{bardes_vicreg_2022} methods, due to their close connection with the information-theoretic framework \cite{shwartz-ziv_information-theoretic_2023} and their underlying assumption about data distribution. 

\paragraph{Probabilistic embeddings (PEs)}
PEs predict a distribution of embeddings rather than a point estimate vector, providing benefits like stable training on noisy data, accurate aggregation and allowing out-of-distribution detection using predicted uncertainty \cite{chun_probabilistic_2021}. They have found applications in diverse areas like face and word embeddings, particularly where understanding representation uncertainty or its intrinsic hierarchies is key \cite{karpukhin_probabilistic_2022}. In addition to their standalone utility, they are often combined with contrastive loss methods \cite{oh_modeling_2019, shi_probabilistic_2019, chun_probabilistic_2021}. In the context of self-supervised learning, Prob-CLR \cite{xie_prob-clr_2021} integrates PEs into the SimCLR architecture, enhancing its clustering capabilities, but makes assumptions that simplify objectives at the cost of uncertainty estimation. Another work \cite{kirchhof_probabilistic_2023} introduces a probabilistic variant of the InfoNCE method, which captures the true generative process posteriors and their underlying variances, offering a more human-aligned understanding of uncertainty. Motivated by these results, we investigate applying PEs to Barlow Twins and VICReg models.

\paragraph{Out-of-distribution (OOD) detection in SSL}
Self-supervised learning methods have shown the capacity to enhance robustness and uncertainty estimation. \cite{ardeshir_uncertainty_2022} successfully leverages features derived from a contrastively learned model to identify OOD examples. Other studies have explored directions such as combining self-supervised with supervised learning objective \cite{winkens_contrastive_2020, hendrycks_using_2019}, using hard data augmentations to push away samples \cite{tack_csi_2020} or incorporating probabilistic modeling to derive uncertainty estimates \cite{kirchhof_probabilistic_2023, nakamura_representation_2023}. In particular, \cite{kirchhof_probabilistic_2023} models the embeddings with von Mises Fischer distribution, where the concentration parameter serves as an uncertainty metric. Similarly, \cite{nakamura_representation_2023} examines SimSiam \cite{chen_exploring_2020} within the variational inference framework and utilizes a power spherical distribution to characterize the distribution of embeddings.
In our work, we employ a similar methodology, utilizing the variance of embeddings as a measure of uncertainty.

\paragraph{Representation vs. loss space}
In self-supervised learning, a common practice is to utilize a non-linear projection head $g(h)$, a neural network component that maps the feature representations ($H$) to a space ($Z$) where the contrastive loss can be optimized more effectively (see Image \ref{fig:bayes_ssl_diagram}).
Our study highlights the vital distinction between representation and loss space. \cite{gupta_understanding_2022} showed that the projection head serves as a low-rank mapping, identifying and mapping certain features to optimize contrastive loss. This is supported by \cite{jing_understanding_2022}, who found the nonlinear projection head filters out discriminative features. Layers closer to the loss space lose more information, hindering generalization. \cite{bordes_guillotine_2023} revealed that misalignment between self-supervised learning and classification tasks is a primary cause for certain phenomena.

\section{Information-theoretic background}
The information-theoretic perspective provides essential insights into the underlying mechanics of self-supervised learning. While certain models, such as InfoMax \cite{linsker_self-organization_1988} and Deep InfoMax \cite{hjelm_learning_2019}, advocate for maximal data information capture, the Information Bottleneck (IB) principle calls for a delicate balance between informativeness and compression \cite{tishby_information_2000, alemi_deep_2019}. 
Our work draws from the multi-view information bottleneck framework \cite{federici_learning_2020}, aiming to capture the essential predictive information shared across different data views.

\textbf{Barlow Twins} This technique can be seen as an IB method, where we aim to maximize the information between the image and representation while minimizing the information about data augmentation, essentially making the representation invariant to distortions. This objective can be represented by the following equation:
\begin{equation}
\begin{split}
\mathcal{L}& = I(Z; V) - \beta I(Z; X) 
\\ & = [\mathcal{H}(Z) - \mathcal{H}(Z | V)]-\beta[\mathcal{H}(Z)-\mathcal{H}(Z | X)]
\\ & = \mathcal{H}(Z | X)+\frac{1-\beta}{\beta} \mathcal{H}(Z)
\\ & = \mathbb{E}_x [\log |\Sigma_{z | x}|]+\frac{1-\beta}{\beta} \log |\Sigma_{z}|,
\end{split}
\end{equation}
where, $X$, $V$ and $Z$ represent original images, augmented views and embeddings, respectively, while $I(\cdot;\cdot)$ and $\mathcal{H}$ denote the mutual information and the entropy. If we assume a Gaussian distribution for the embeddings, the entropy terms within the objective can be reduced to the log-determinant of their corresponding covariance functions. 
Notably, Barlow Twins does not optimize covariance matrices directly but instead uses a proxy objective (see Section \ref{sec:ssl_framework}). Additionally, the IB formulation in this context doesn't directly address the multi-view characteristic intrinsic to self-supervised learning.

\textbf{VICReg} This method, on the other hand, can be contextualized within the multi-view perspective \cite{shwartz-ziv_information-theoretic_2023}, and it is possible to derive the VICReg objective from an information-theoretic standpoint, leveraging a lower bound derived from \cite{federici_learning_2020}. It can be expressed as maximizing the information between the views and their corresponding embeddings:
\begin{equation}
\begin{split}
\text{max.}\: I(Z, V^{\prime})&=\mathcal{H}(Z)-\mathcal{H}(Z | V^{\prime}) \\ & \geq \mathcal{H}(Z) + \mathbb{E}_{v,z|v,v^\prime,z^\prime|v^\prime}[\log q(z|z^\prime)]
\end{split}
\end{equation}
Notably, the most challenging aspect of this perspective lies in estimating the entropy term $\mathcal{H}(Z)$, which is generally computationally intractable. Again, the covariance and variance terms within the VICReg framework serve as proxies for maximizing the entropy term. This can be achieved by diagonalizing the covariance matrix, e.g., by increasing values along its diagonal while pushing the off-diagonal terms towards zero.


\section{Methodology}
\subsection{Self-supervised learning framework}
\label{sec:ssl_framework}
First, let us formalize the feature decorrelation-based SSL framework setup for point estimate embeddings.
We sample an image $x$ from a dataset $\mathcal{D}$ and create two views $v$ and $v^\prime$ by applying transforms $t$ and $t^\prime$ sampled from a distribution $\mathcal{T}$. These views are then fed into an encoder $f_\theta$, parameterized by $\theta$, to create representations $h=f_\theta(v)$ and $h^\prime=f_\theta(v^\prime)$. Next, the representation vectors are passed through the projector $g_\phi$, parameterized by $\phi$, to obtain embeddings $z=g_\phi(h)$ and $z^\prime=g_\phi(h^\prime)$. The loss function $\mathcal{L}(\cdot, \cdot)$ is then applied to these embeddings $z$ and $z^\prime$. 

\begin{figure}[h]
    \centering
    \includegraphics[width=0.49\textwidth]{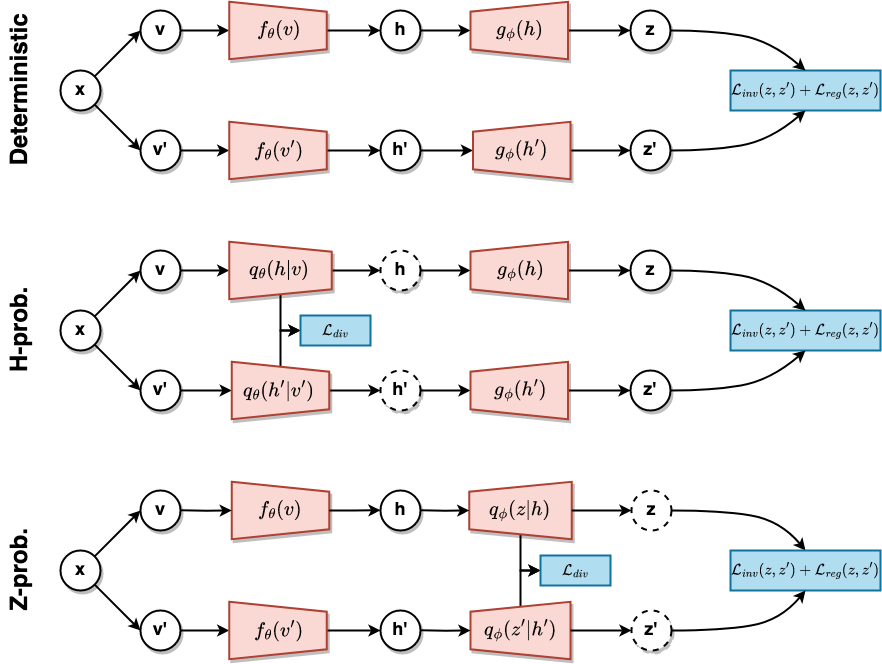}
    \caption{Diagram of three approaches: (1) \textbf{Deterministic}, with direct mappings using \(f_\theta\) and \(g_\phi\), (2) \textbf{H-prob}, introducing stochasticity with \(q_\theta\) before direct mapping \(g_\phi\), and (3) \textbf{Z-prob}, with probabilistic encoding \(q_\phi\) post-deterministic mapping \(f_\theta\). Losses \(\mathcal{L}_{\text{inv}}(z, z')\) and \( \mathcal{L}_{\text{reg}}(z, z')\) are computed using the embeddings. \(\mathcal{L}_{\text{div}}\) loss is the KL divergence with a predefined prior, e.g., standard normal distribution $\mathcal{N}(0,1)$}.
    \label{fig:bayes_ssl_diagram_specific}
\end{figure}

In general, we can define the loss function for these methods as:
\begin{equation}
\label{eq:ssl_general_loss}
\mathcal{L}(Z, Z^\prime) = \mathcal{L}_{inv}(Z, Z^\prime) + \mathcal{L}_{reg}(Z, Z^\prime),
\end{equation}
where $Z = [z_1, \ldots, z_n]$ and $Z^\prime = [z^\prime_1, \ldots, z^\prime_n]$ represent a batch of $n$ embedding vectors of dimension $d$ for sampled image views.

\paragraph{Barlow Twins} In this method, the loss function is computed using the cross-correlation matrix $R$ on embeddings, which are mean-centred along the batch dimension:
\begin{equation}
\begin{split}
R_{ij} &= \text{corr}(\sum_b Z_{bi}, \sum_b Z^\prime_{bj}) \\ & = \frac{\text{cov}(\sum_b Z_{bi}, \sum_b Z^\prime_{bj})}{\sigma_{\sum_b Z_{bi}} \sigma_{\sum_b Z^\prime_{bj}}},
\end{split}  
\end{equation}

where $b$ indexes batch samples, $i$ and $j$ index the embedding vector dimensions, and $\sigma$ denotes standard deviation. From this matrix, we compute the invariance term:
\begin{equation}
  \mathcal{L}_{inv}(Z, Z^\prime) = \sum_i(1-R_{i i})^2, 
\end{equation}
and regularization (feature decorrelation) term
\begin{equation}
\mathcal{L}_{reg}(Z, Z^\prime) = \lambda \sum_i \sum_{j \neq i} R_{i j}^2,
\end{equation}
 where $\lambda$ is a loss scaling coefficient.
 
\paragraph{VICReg} In contrast, VICReg calculates the decorrelation term from a covariance matrix:
\begin{equation}
    \mathcal{L}_{cov}(Z) = \frac{1}{d} \sum_i \sum_{j \neq i} C_{ij},
\end{equation}
which is simply the sum of the squared off-diagonal coefficients of the covariance matrix $C_{ij} = \text{cov}(\sum_b Z_{bi}, \sum_b Z^\prime_{bj})$.

To prevent collapse, VICReg includes a variance regularization term to the loss:
\begin{equation}
\mathcal{L}_{var}(Z) = \frac{1}{d} \sum_{i=1}^d \max(0, \gamma-\sigma_{\sum_b Z_{bi} + \epsilon})
\end{equation}
which is a hinge function that operates on the standard deviation of the embeddings across the batch dimension. VICReg utilizes $\gamma = 1$ and a small $\epsilon$ value to avoid numerical instabilities.

Both regularization terms are calculated separately for $Z$ and $Z^\prime$ using $\tau$ and $\nu$ as loss coefficient:
\begin{equation}
\begin{split}
\mathcal{L}_{reg}(Z, Z^\prime) &=
\tau [\mathcal{L}_{var}(Z) + \mathcal{L}_{var}(Z^\prime)]
\\ & + \nu [\mathcal{L}_{cov}(Z) + \mathcal{L}_{cov}(Z^\prime)]
\end{split}  
\end{equation}
The invariance term is calculated using mean-squared error loss with $\alpha$ as a loss coefficient:
\begin{equation}
\mathcal{L}_{inv}(Z, Z^\prime) = \frac{\alpha}{n} \sum_{i=1}^n\|Z_i-Z_i^{\prime}\|_2^2.
\end{equation}

\subsection{Probabilistic embeddings}
\label{sec:probabilistic_embeddings}
We propose to extend the aforementioned SSL frameworks to probabilistic embeddings. Drawing inspiration from the works of \cite{federici_learning_2020} and \cite{shwartz-ziv_information-theoretic_2023}, we can reformulate our self-supervised objective as an information maximization problem. 
We aim to maximize the mutual information between the views and their corresponding embeddings, i.e., $I(Z; V^\prime)$ and $I(Z^\prime; V)$. We utilize the following lower bound:
\begin{equation}
\label{eq:mv_ib_vicreg}
\begin{split}
I(Z; V^\prime) &= \mathcal{H}(Z) - \mathcal{H}(Z|V^\prime) \\ & \geq \mathcal{H}(Z) + \mathbb{E}_{v^\prime}[\log q(z|v^\prime)]
\\ & \geq \mathcal{H}(Z) + \mathbb{E}_{z|v}[\mathbb{E}_{z^\prime|v^\prime}[\log q(z|z^\prime)]]
\end{split}
\end{equation}

From \cite{shwartz-ziv_information-theoretic_2023}, we know that the first term in this lower bound, $\mathcal{H}(Z)$, is implicitly optimized by our regularization term $\mathcal{L}_{reg}$. Furthermore, we address the second term in this lower bound by optimizing the invariance term $\mathcal{L}_{inv}$. Therefore, we recover the objective from Eq. \ref{eq:ssl_general_loss}.
Expectations from Eq. \ref{eq:mv_ib_vicreg} are evaluated over empirical data distribution. Specifically, we obtain the expectations by backpropagating through $K$ Monte Carlo (MC) samples using the reparametrization trick \cite{kingma_auto-encoding_2013}:
\begin{multline}
\label{eq:mc_mv_ib_vicreg}
    \mathbb{E}_{z|v}[\mathbb{E}_{z^\prime|v^\prime}[\log q(z|z^\prime)]] \simeq \\ \frac{1}{nK}\sum_{i=1}^n\sum_{k=1}^K \log q(z_{ik}|z^\prime_{ik}).
\end{multline}

Similarly to Eq. \ref{eq:mc_mv_ib_vicreg}, we estimate the expected value of the regularization loss by evaluating the posteriors according to the specific model.

\paragraph{Stochastic loss space (Z-prob.)}
In this model variant, we first deterministically encode the image view $v$ into the representation $h$ using a deterministic encoder $f_\theta$. Then we sample $z$ based on $h$ using the stochastic projector $q_\phi(z|h)$ (which follows a Normal distribution). For ease of notation, we use $q'(z|h)$ defined as follows:

\begin{equation}
q'(z|v) = q_\phi(z|f_\theta(v)), \;
z \sim \mathcal{N}\left(z | \mu_\phi(h), \sigma_\phi^2(h)I\right)
\end{equation}

We apply the same procedure for the second image view $v^\prime$ to produce the representation $h^\prime$ and the embedding $z^\prime$, utilizing the same encoder and projector parameters $\theta$ and $\phi$.

\paragraph{Stochastic representation space (H-prob.)}
When considering probabilistic embeddings within the representation space $H$, our derivation must also account for the presence of $h$. Let's assume the following joint distribution of $q(v, h, z)$:
$$q(v, h, z) = q(z|v, h) q(h|v) q(v).$$
Given that $z$ depends on $h$, computing $q(z|v)$ necessitates marginalizing out $h$.
Consequently, the expectation term from Eq. \ref{eq:mv_ib_vicreg} can be expressed as:
\begin{equation}
\mathbb{E}_{v^\prime}[\log q(z|v^\prime)] \doteq  \mathbb{E}_{h, v^\prime}[\log q(z|h, v^\prime)].
\end{equation}

We sample $h$ using the stochastic encoder $q_\theta(h|v)$, which follows a Normal distribution:
\begin{equation}
h \sim \mathcal{N}(h | \mu_\theta(v), \sigma_\theta^2(v)I)
\end{equation}

Then we obtain the embedding $z$ by mapping the representation $h$ with a projector, $g_\phi$.
We apply the same procedure for the second image view $v^\prime$ to produce the representation $h^\prime$ and the embedding $z^\prime$, utilizing the same encoder and projector parameters $\theta$ and $\phi$.

\paragraph{Regularization}
Moving from point estimates to probabilistic embeddings, we introduce an additional layer of uncertainty, which helps capture the inherent ambiguity and variability in the data. However, it also raises the challenge of regularizing this stochasticity to prevent trivial solutions and obtain reliable uncertainty estimates.
To address this issue, we follow the Variational Information Bottleneck (VIB) framework \cite{alemi_deep_2019} and formulate an additional regularization term to the loss function in the form of a KL divergence between the probabilistic embeddings $q(\cdot|v)$ and $q(\cdot|v')$, and a predefined prior $\hat{q}(\cdot)$, typically $\mathcal{N}(0, 1)$:
\begin{equation}
\mathcal{L}_{div} = \frac{\beta}{2} [\text{KL}(q(\cdot|v) || \hat{q}(\cdot)) + \text{KL}(q(\cdot|v') || \hat{q}(\cdot))],   
\end{equation}
This regularization, controlled by $/beta$ parameter, acts as a bottleneck, constraining the capacity of our probabilistic embeddings, which has been shown to be effective in previous work \cite{achille_emergence_2018, alemi_deep_2019} in terms of improving robustness and disentanglement. 
The overall modified loss function becomes:
\begin{equation}
\mathcal{L} = \mathcal{L}_{inv} + \mathcal{L}_{reg} + \mathcal{L}_{div}.   
\end{equation}


\section{Experiments}
\label{sec:experiments}

\subsection{Setup}
\label{sec:experimental_setup}
We pre-train our model in a self-supervised manner (without labels) on the ImageNet ILSVRC-2012 dataset \cite{russakovsky_imagenet_2015}.
We adopt the same image augmentations and closely adhere to the original works in determining the loss coefficients. 
We opt for the smaller ResNet-18 \cite{he_deep_2015} architecture as our backbone encoder and a smaller projector. Our experimental setup involves training the model for 100 epochs with a batch size of 256 using AdamW \cite{loshchilov_decoupled_2019} optimizer. We rely on an $\mathcal{N}(0, 1)$ prior for Barlow Twins and VICReg methods across both models (H- and Z-prob.) and use 12 Monte Carlo (MC) samples. For more details, see Appendix \ref{sec:app_pretraining}. A more comprehensive exploration of PEs hyperparameters is conducted in the ablation study (Section \ref{sec:ablations}).

\subsection{ImageNet evaluation}
We follow the evaluation procedure of \cite{goyal_scaling_2019}, as laid out in the original works of Barlow Twins and VICReg and report the results for linear classification and semi-supervised learning tasks.
For the linear classification, we train a linear classifier on the frozen representation from our pre-trained backbone encoder. Importantly, for the $\text{H-prob.}$ method, which yields an embedding distribution, we compute the final representation as an average over posterior samples (see: Eq. \ref{eq:mc_mv_ib_vicreg}). 
For the semi-supervised learning task, both the backbone encoder and a linear classifier are fine-tuned. We utilize ImageNet subsets corresponding to 1\% and 10\% of the labels \cite{chen_exploring_2020}. A detailed procedure can be found in the Appendix \ref{sec:app_imagenet_evaluation}. Our training process was conducted once due to computational constraints and training stability \cite{bardes_vicreg_2022}.

\begin{table}[htb]
    \centering
    \caption{Comparison of top-1 accuracies for both linear and semi-supervised tasks on ImageNet.}
    \label{tab:imagenet}
    \begin{adjustbox}{width=0.5\textwidth}
    \begin{tabular}{@{}llrrrr@{}}
\toprule
                              &               & \multicolumn{2}{l}{Linear} & \multicolumn{2}{l}{Semi-supervised} \\ \midrule
Method                        & Embeddings    & \multicolumn{2}{l}{}       & 1\%              & 10\%             \\ \midrule
\multirow{3}{*}{VICReg}       & Deterministic & 0.490          &           & 0.315            & 0.509            \\
                              & Z-prob.       & 0.484          &           & 0.310            & 0.507            \\
                              & H-prob.       & 0.454          &           & 0.313            & 0.498            \\ \midrule
\multirow{3}{*}{Barlow Twins} & Deterministic & 0.495          &           & 0.316            & 0.510            \\
                              & Z-prob.       & 0.489          &           & 0.313            & 0.506            \\
                              & H-prob.       & 0.451          &           & 0.309            & 0.495            \\ \bottomrule
\end{tabular}
    \end{adjustbox}
\end{table}

The results are presented in Table \ref{tab:imagenet}. From our observations, the probabilistic embeddings in the loss space (Z-prob.) generally outperform the probabilistic embeddings in the representation space (H-prob.) for both the Barlow Twins and VICReg methods, particularly in the linear classification task. This phenomenon is further explained in Section \ref{sec:info_compression}, where we demonstrate that this distinction arises due to the significant amount of information shared between the representation and loss spaces. It is noteworthy that the difference becomes lower in the semi-supervised task, especially for 1\% of available labels.

\subsection{Transfer learning}
To further verify the implications of probabilistic embeddings, we perform transfer learning experiments  \cite{zbontar_barlow_2021, bardes_vicreg_2022}. 
Specifically, we freeze the encoder pre-trained on ImageNet and train a single linear layer on top of it. 
We compare probabilistic embeddings (H- and Z-prob.) to their deterministic counterparts for Barlow Twins and VICReg methods. To this end, we utilize three datasets for evaluation: INaturalist \cite{van_horn_inaturalist_2018}, SUN397 \cite{xiao_sun_2010}, and Flowers-102 \cite{nilsback_automated_2008}. The detailed hyperparameters are in Appendix \ref{sec:app_transfer_learning}. The results are shown in Table \ref{tab:transfer_learning}. As in the previous experiments, $\text{Z-prob.}$ embeddings yield results on par with the Deterministic approach, while for $\text{H-prob.}$ we observe a performance degradation.

\begin{table}[htb]
    \centering
    \caption{Comparison of top-1 accuracy on transfer learning experiments.}
    \label{tab:transfer_learning}
    \begin{adjustbox}{width=0.5\textwidth}
    \begin{tabular}{@{}llrrr@{}}
\toprule
\multicolumn{1}{l}{Method}    & \multicolumn{1}{l}{Embeddings} & \multicolumn{1}{c}{INat18} & \multicolumn{1}{c}{SUN397} & \multicolumn{1}{c}{Flowers-102}   \\ 
\midrule
\multirow{3}{*}{VICReg}     & Deterministic            & 0.154 & 0.477 &  0.649 \\   
                            & Z-prob.                  & 0.147 & 0.478 & 0.629 \\
                            & H-prob.                  & 0.110 & 0.463 & 0.617 \\
\midrule
\multirow{3}{*}{Barlow Twins}   & Deterministic            & 0.148 & 0.482 & 0.645 \\
                                & Z-prob.                  & 0.149 & 0.481 & 0.645 \\
                                & H-prob.                  & 0.126 & 0.458 & 0.626 \\
\midrule
\end{tabular}
    \end{adjustbox}
\end{table}


\subsection{Out-of-distribution detection}
\label{sec:ood}
To investigate the out-of-distribution capabilities of our probabilistic embeddings, we follow a similar evaluation procedure to \cite{liang_enhancing_2020} and train our models on the CIFAR-10 dataset \cite{krizhevsky_learning_2009}. We consider the original test set of CIFAR-10 as IN data and assess its ability to distinguish between other OUT datasets, such as Textures \cite{cimpoi_describing_2013}, TinyImageNet(crop, resized) \cite{le_tiny_2015} and LSUN(crop, resized) \cite{yu_lsun_2016}.
We introduce detectors based on the first two moments of the embeddings' variance (denoted as SigmaMean and SigmaStd) and compare them to other detectors.  We report the averaged AUROC metric over three runs and all OUT datasets -- see Table \ref{tab:ood_main_max}.

As observed, leveraging intrinsic properties of stochastic embeddings, such as their variance, can be highly effective as an OOD detector. In some instances, it matches or surpasses the performance of detectors relying on label information. More OOD experiment details can be found in Appendix \ref{app:ood}.

\begin{table}[h]
\begin{center}
\caption{AUROC performance for the OOD detection. Entropy \cite{chan_entropy_2021}, MaxSoftmax \cite{hendrycks_baseline_2018} and ODIN \cite{liang_enhancing_2020} require label information, whereas Mahalanobis \cite{lee_simple_2018}, SigmaMean, and SigmaStd utilize only representations.}
\begin{adjustbox}{width=0.5\textwidth}
\begin{tabular}{@{}llrrr@{}}
\toprule
Method                        & Detector    & Z-prob. & H-prob. & Deterministic \\ \midrule
\multirow{6}{*}{VICReg}       & Entropy     & 0.827   & 0.832   & 0.827         \\
                              & MaxSoftmax   & 0.801   & 0.805   & 0.799         \\
                              & ODIN      & 0.793   & 0.793   & 0.788         \\ \cmidrule(l){2-5} 
                              & Mahalanobis & 0.689   & 0.735   & 0.649         \\
                              & SigmaMean   & 0.788   & 0.768   & \text{N/A}             \\
                              & SigmaStd    & 0.781   & 0.707   & \text{N/A}             \\ \midrule
\multirow{6}{*}{Barlow Twins} & Entropy     & 0.831   & 0.828   & 0.823         \\
                              & MaxSoftmax  & 0.803   & 0.798   & 0.795         \\
                              & ODIN        & 0.794   & 0.797   & 0.783         \\ \cmidrule(l){2-5} 
                              & Mahalanobis & 0.719   & 0.681   & 0.654         \\
                              & SigmaMean   & 0.691   & 0.749   & \text{N/A}             \\
                              & SigmaStd    & 0.831   & 0.795   & \text{N/A}             \\ \midrule
\end{tabular}
\label{tab:ood_main_max}
\end{adjustbox}
\end{center}
\end{table}


\section{Ablations}
\label{sec:ablations}
The ablation study was performed on the CIFAR-10 dataset. Each model was trained on three different seeds for 200 epochs with a batch size of 256. Appendix \ref{app:ablation} provides a more detailed setup for the ablation study.

\paragraph{Prior}
We compare $\mathcal{N}(0, 1)$ prior to a Mixture of Gaussians (MoG) to study the effect of using a more expressive distribution for modeling our probabilistic embeddings. The MoG prior has the following form: $\frac{1}{M} \sum_{m=1}^M \mathcal{N}\left(\mu_m, \operatorname{diag}\left(\sigma_m^2\right)\right),$
where $M$ denotes the number of mixtures, while $\mu_m$ and $\sigma_m$ denote trainable parameters of a specific Gaussian in the mixture model. The results are shown in Table \ref{tab:ablation_prior}. We can see that contrary to our intuition, the effect of MoG on performance is insignificant, often degrading the model's efficacy.  

\begin{table}[h]
\caption{Comparison of different priors effect on model's performance measured by top-1 accuracy.}
\begin{center}
\begin{adjustbox}{width=0.5\textwidth}
\begin{tabular}{@{}llrlrr@{}}
\toprule
Method                        & Embeddings               & Beta ($\beta$)  &  & MoG   & $\mathcal{N}(0, 1)$ \\ \midrule
\multirow{6}{*}{Barlow Twins} & \multirow{3}{*}{H-prob.} & 1e-04 &  & 0.798 & 0.802  \\
                              &                          & 1e-03 &  & 0.793 & 0.804  \\
                              &                          & 1e-02 &  & 0.795 & 0.797  \\ \cmidrule(l){2-6} 
                              & \multirow{3}{*}{Z-prob.} & 1e-03 &  & 0.823 & 0.824  \\
                              &                          & 1e-02 &  & 0.819 & 0.827  \\
                              &                          & 1e-01 &  & 0.747 & 0.82   \\ \midrule
\multirow{6}{*}{VICReg}       & \multirow{3}{*}{H-prob.} & 1e-05 &  & 0.824 & 0.827  \\
                              &                          & 1e-04 &  & 0.825 & 0.826  \\
                              &                          & 1e-03 &  & 0.806 & 0.811  \\ \cmidrule(l){2-6} 
                              & \multirow{3}{*}{Z-prob.} & 1e-05 &  & 0.833 & 0.833  \\
                              &                          & 1e-04 &  & 0.831 & 0.833  \\
                              &                          & 1e-03 &  & 0.793 & 0.825  \\ \bottomrule
\end{tabular}
\label{tab:ablation_prior}
\end{adjustbox}
\end{center}
\end{table}

\paragraph{Beta scale}
We study the effect of different $\beta$ scales on model performance. As mentioned in Section \ref{sec:probabilistic_embeddings}, this term controls the bottleneck and, therefore, the capacity of the embeddings. The results are presented in Table \ref{tab:ablation_beta}. We observe better model performance for sufficiently small $\beta$, while higher values may deteriorate the model's efficacy. However, by reducing $\beta$ too much, the variance of the embeddings reduces accordingly, making the embeddings \textit{more deterministic}, as shown in Appendix \ref{app:probabilistic_embeddings}.

\begin{table}[h]
\caption{The effect of beta ($\beta$) on model's performance (top-1 accuracy). The "-" indicates that a particular $\beta$ value was not explored for a given method due to differences in loss scales between the models, necessitating distinct sets of $\beta$ values for each method.}
\begin{center}
\begin{adjustbox}{width=0.5\textwidth}
\begin{tabular}{llrrrrr}
\hline
Method                        & Embeddings & 1e-05 & 1e-04 & 1e-03 & 1e-02 & 1e-01 \\ \hline
\multirow{2}{*}{Barlow Twins} & H-prob.    & -     & 0.802 & 0.804 & 0.797 & -     \\
                              & Z-prob.    & -     & -     & 0.824 & 0.827 & 0.82  \\
\multirow{2}{*}{VICReg}       & H-prob.    & 0.827 & 0.826 & 0.811 & -     & -     \\
                              & Z-prob.    & 0.833 & 0.833 & 0.825 & -     & -     \\ \hline
\end{tabular}
\label{tab:ablation_beta}
\end{adjustbox}
\end{center}
\end{table}

\paragraph{Number of Monte Carlo samples}
In this experiment, we study the benefit of using multiple MC samples to estimate the expectation from Eq. \ref{eq:mc_mv_ib_vicreg}. Following the VIB framework, we use either 1 or 12 samples. The results of this experiment are presented in Table \ref{tab:ablation_mc_samples}. 
For the $\text{H-prob.}$ embeddings, increasing the number of MC samples improves model performance, particularly with higher values of $\beta$. This indicates that using more MC samples offers a better and less biased estimation of the expectations.  
Conversely, the number of MC samples seems to have an insignificant effect on the $\text{Z-prob.}$ embeddings.

\begin{table}[h]
\caption{The effect of number of MC samples on model's performance measured by top-1 accuracy.}
\begin{center}
\begin{adjustbox}{width=0.5\textwidth}
\begin{tabular}{llrlrr}
\hline
Method                        & Embeddings               & Beta ($\beta$)  &  & 1     & 12    \\ \hline
\multirow{6}{*}{Barlow Twins} & \multirow{3}{*}{H-prob.} & 1e-04 &  & 0.801 & 0.803 \\
                              &                          & 1e-03 &  & 0.799 & 0.809 \\
                              &                          & 1e-02 &  & 0.790  & 0.805 \\ \cline{2-6} 
                              & \multirow{3}{*}{Z-prob.} & 1e-03 &  & 0.823 & 0.826 \\
                              &                          & 1e-02 &  & 0.827 & 0.827 \\
                              &                          & 1e-01 &  & 0.821 & 0.819 \\ \hline
\multirow{6}{*}{VICReg}       & \multirow{3}{*}{H-prob.} & 1e-05 &  & 0.826 & 0.829 \\
                              &                          & 1e-04 &  & 0.823 & 0.828 \\
                              &                          & 1e-03 &  & 0.804 & 0.817 \\ \cline{2-6} 
                              & \multirow{3}{*}{Z-prob.} & 1e-05 &  & 0.834 & 0.831 \\
                              &                          & 1e-04 &  & 0.834 & 0.831 \\
                              &                          & 1e-03 &  & 0.824 & 0.826 \\ \hline
\end{tabular}
\label{tab:ablation_mc_samples}
\end{adjustbox}
\end{center}
\end{table}


\section{Information compression}
\label{sec:info_compression}
In this section, we analyze the information compression factor of probabilistic embeddings. As they introduce an additional bottleneck to the network, we aim to explore its real effect on the amount of information conveyed between these pairs of spaces: image views ($V$), representations ($H$) and embeddings ($Z$). 
To this end, we employ Mutual Information Neural Estimation (MINE) \cite{belghazi_mine_2021}, which provides an efficient and scalable way to estimate mutual information between the aforementioned spaces.

\begin{figure}[h]
    \centering
    \includegraphics[width=0.45\textwidth]{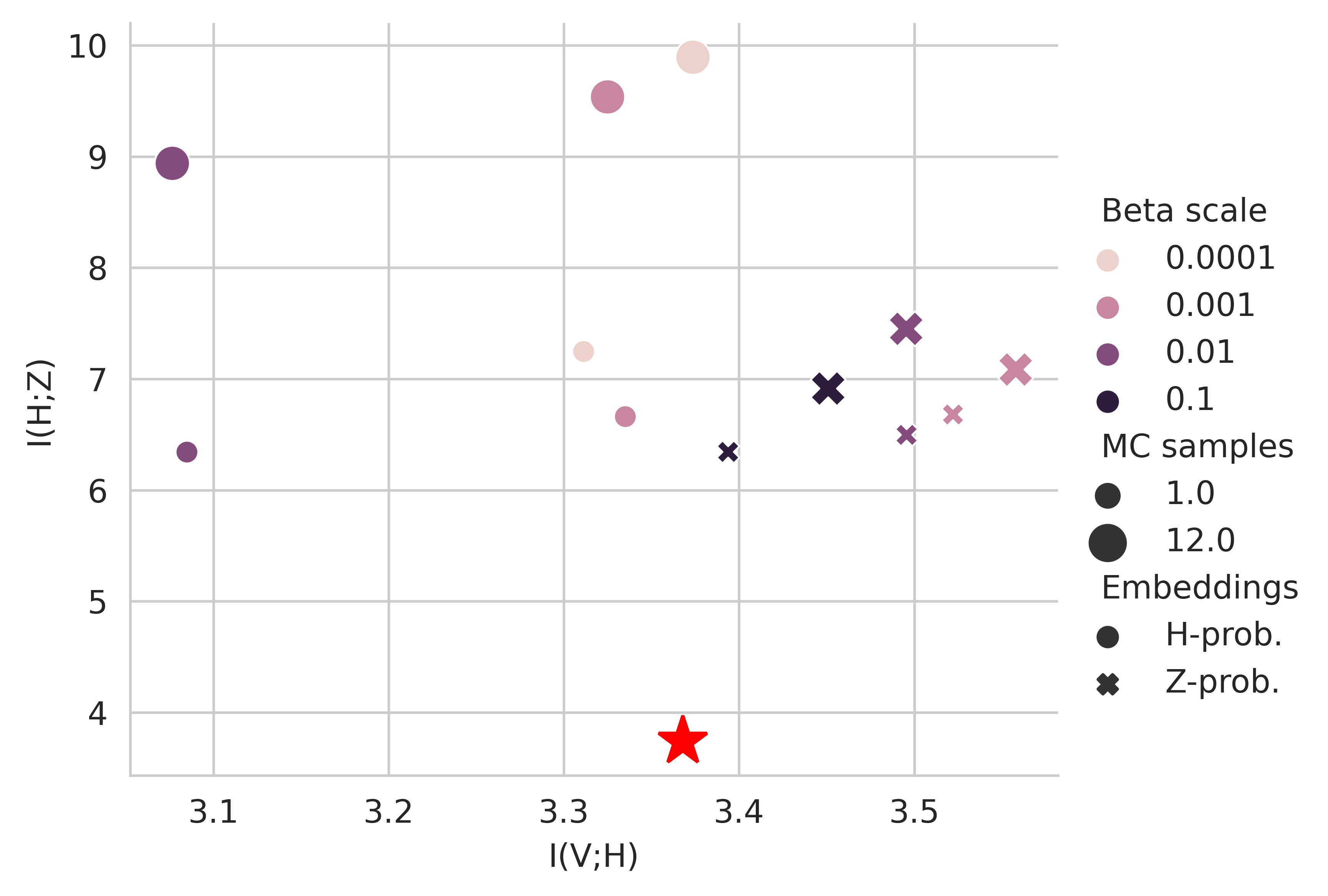}
    \caption{$I(V;H)$ vs. $I(H;Z)$ for Barlow Twins method. We can see that the amount of information shared between representation and loss space varies drastically between different models. Red star denotes the deterministic (original) method.}
\label{fig:information_compression_VH_HZ}
\end{figure}

The relationship between the mutual information $I(V; H)$ and $I(H; Z)$ for the Barlow Twins method is illustrated in Figure \ref{fig:information_compression_VH_HZ}.

We hypothesize that the reason for the weaker performance of $\text{H-prob.}$ embeddings is high mutual information between representation and loss space $I(H;Z)$. 
We believe that is due to the constrained representation, containing more discriminative features relevant to the contrastive loss, losing generic information leveraged by a downstream task. Nonetheless, $\text{Z-prob.}$ embeddings contain more information about the image views $I(V;H)$, which helps them mitigate such effect (see \cite{wang_rethinking_2022}).

Furthermore, we observe that using more MC samples enhances the model's performance for higher values of $\beta$. We attribute this improvement to the increased amount of information shared between the two views, $z$ and $z^\prime$.
According to \cite{federici_learning_2020}, this information is the predictive information of representation, which corresponds to the invariance term in the SSL loss function (see Section \ref{sec:probabilistic_embeddings}).  

\begin{figure}[h]
    \centering
    \includegraphics[width=0.49\textwidth]{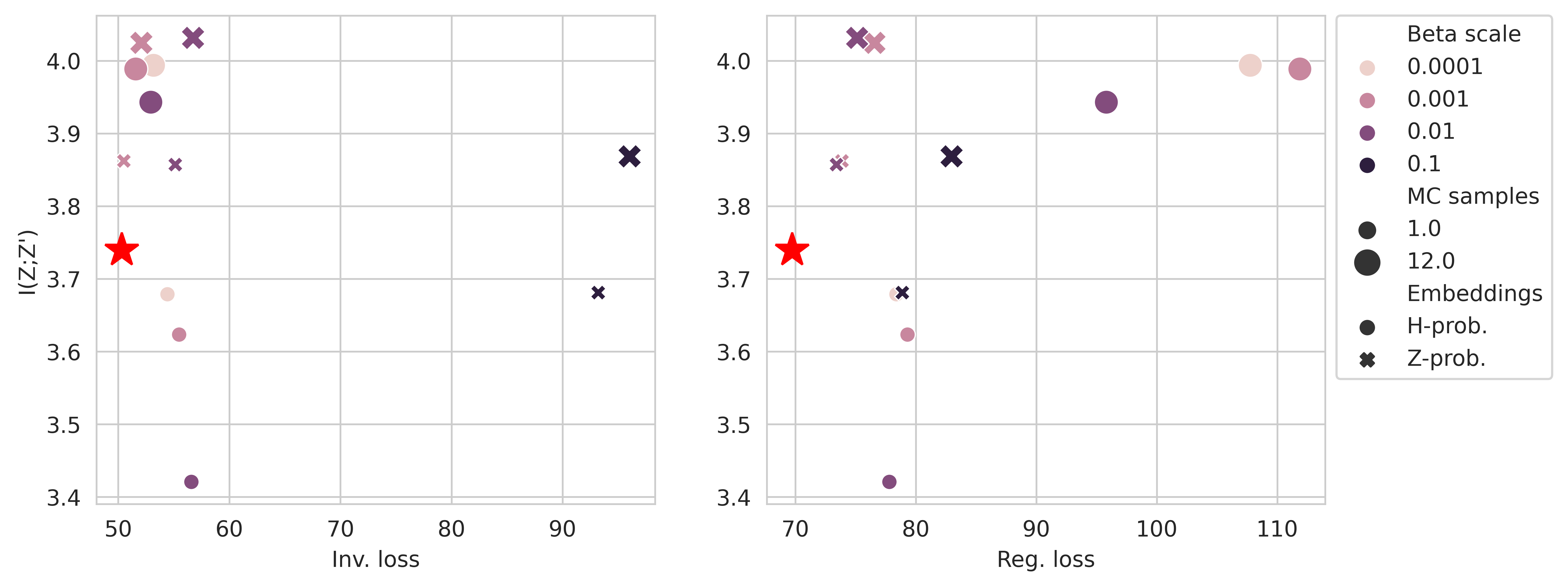}
    \caption{Information shared between views $I(Z;Z')$  vs. loss terms in Barlow Twins.}
    \label{fig:information_compression_ZZ_losses}
\end{figure}

Figure \ref{fig:information_compression_ZZ_losses} shows that for a higher number of MC samples, the information $I(Z;Z^\prime)$ increases and invariance loss decreases accordingly. Even though the MC samples exacerbate the regularization loss, we can improve the model's performance. The extensive results regarding the mutual information are given in Appendix \ref{app:information_bottleneck}. 

\section{Conclusions}
In this work, we unveiled the potential of probabilistic embeddings in self-supervised learning. We introduced probabilistic embeddings to feature decorrelation-based methods, specifically Barlow Twins and VICReg. When applied in the loss space ($\text{Z-prob.}$), these embeddings performed on par with the deterministic counterparts. However, their incorporation into the representation space ($\text{H-prob.}$), posed challenges, leading to compromised downstream performance.
We performed a thorough analysis of mutual information in the network and posit that this is due to an overemphasis on data invariance and, in the aftermath, lower generalization. Importantly, our method exhibited a robust ability to detect out-of-distribution samples, even outperforming certain label-based detectors. This showcases the potential of our approach and suggests avenues for future research in self-supervised learning optimization.

\clearpage
\bibliography{main}


\clearpage
\appendix
\section{Detailed procedure and hyperparameters for experiments}
\subsection{Pre-training}
\label{sec:app_pretraining}
In this section, we provide a comprehensive overview of the experimental framework, expanding on the preliminary details shared in Section \ref{sec:experimental_setup}.
We pre-trained our model using a self-supervised approach on the ImageNet ILSVRC-2012 dataset \cite{russakovsky_imagenet_2015}.
In our experiments, we opted for a smaller encoder, ResNet-18, compared to the original Barlow Twins and VICReg studies. Additionally, we used a reduced embedding dimensionality of 1024 and a smaller batch size of 256. We pre-trained the model for a total of 100 epochs. Consequently, we employed a more compact projector network consisting of three fully connected linear layers, each with a dimensionality of 1024, where the initial two layers are followed by batch normalization and ReLU activation. This decision to employ smaller networks and embeddings is primarily driven by the computational constraints we faced during our experiments. However, we posit that our findings can be generalized to larger networks \cite{balestriero_cookbook_2023}. Smaller batch size is driven by our observation that it yields superior results with fewer epochs (based on performed hyperparameter search).
Owing to the unusual behavior exhibited by the SGD optimizer in the self-supervised learning process when using stochastic embeddings, we instead utilized the AdamW \cite{loshchilov_decoupled_2019} optimizer. The SGD exhibited low sensitivity to the $\beta$ scale hyperparameter but higher to the learning rate, e.g., the variance of the embeddings was dictated rather by the learning rate and not the $\beta$ scale. 
We set the learning rate to $1 \times 10^{-3}$ and applied a weight decay of $1 \times 10^{-4}$. Furthermore, we employed a cosine decay scheduler with two warmup epochs, during which the learning rate increases from 0 to $1 \times 10^{-3}$, and then gradually scaled down the learning rate to a final value of $5 \times 10^{-4}$.

\subsection{Linear and semi-supervised evaluation on ImageNet}
\label{sec:app_imagenet_evaluation}
Our experimental approach for linear classification and semi-supervised experiments is similar to methodologies outlined in previous studies \cite{grill_bootstrap_2020, zbontar_barlow_2021, bardes_vicreg_2022}. 
Specifically, for the \textbf{linear classifcation} task, the backbone model (ResNet-18), which was pre-trained under both deterministic and probabilistic conditions, remains fixed while we train a single linear layer atop it. We determined the number of epochs using a validation set. If a validation set was unavailable or utilized as a test set, we executed a stratified split, allocating $20\%$ of the training data. We settled on 100 epochs and chose a batch size of 512. For the optimization of the linear layer, we employed the AdamW optimizer \cite{loshchilov_decoupled_2019}, initiating with a learning rate of $1 \times 10^{-2}$ and decreasing it by a factor of $0.1$ in evenly-spaced epochs: 30, 60 and 90 (until it reaches $1 \times 10^{-5}$). The weight-decay parameter was configured at $1 \times 10^{-4}$. Before passing the representation vector to the linear layer, we performed $L_2$ normalization, aiming to minimize the impact of scale in the evaluated models.
For \textbf{semi-supervised} experiments, we unfroze the backbone model and trained it jointly with the linear layer. For optimization, we utilized the AdamW optimizer. The initial learning rate was set at $1 \times 10^{-3}$ for the linear layer and $1 \times 10^{-4}$ for the backbone, with both rates decreasing by a factor of $0.1$ following a cosine decay schedule. The weight-decay was set to $1 \times 10^{-5}$. The training was conducted with a batch size of 256 across 50 epochs.

\subsection{Transfer Learning} \label{sec:app_transfer_learning}
Similarly to experiments on ImageNet, we have based our experimental protocol for transfer learning experiments on the ones used in previous works \cite{grill_bootstrap_2020, zbontar_barlow_2021, bardes_vicreg_2022}. In particular, the backbone model, pre-trained in deterministic and probabilistic settings, is frozen, and one linear layer is trained on top of it with three datasets: INaturalist \cite{van_horn_inaturalist_2018}, SUN397 \cite{xiao_sun_2010}, and Flowers-102 \cite{nilsback_automated_2008}. We use the same protocol as for linear classification. We leveraged AdamW with learning rate starting from $1 \times 10^{-2}$ and decaying by $0.1$ in evenly-spaced steps to $1 \times 10^{-5}$; weight-decay parameters was set to $1 \times 10^{-4}$. We determined the number of 50, 50, and 80 epochs for INaturalist, SUN397, and Flowers-102, respectively. As for batch size, we used 256 for INaturalist and SUN397 and 128 for Flowers-102. 

\subsection{Ablation study}
\label{app:ablation}
In terms of the model's architecture, our configuration during ablation studies on CIFAR-10 mirrors that of our principal ImageNet experiments. Specifically, we employ a ResNet-18 encoder as a backbone, complemented by a non-linear projection head. This head comprises three fully connected linear layers, each featuring a dimensionality of 1024. Notably, the first two of these layers incorporate batch normalization and ReLU activation functions.
For the optimization process, we adopted the AdamW optimizer with a weight decay of $1 \times 10^{-4}$. Our initial learning rate was set at $1 \times 10^{-3}$ with a cosine decay scheduling, which gradually scaled down the learning rate to a final value of $1 \times 10^{-5}$. We set the batch size to 256 and trained the model on three different seeds for 200 epochs. The variability in the loss function's magnitude and method-specific sensitivities necessitated the selection of distinct beta ($\beta$) scale hyperparameters for each approach, as documented in Table \ref{tab:ablation_beta}. 
In the \textbf{prior} ablation study, we reported results (Table \ref{tab:ablation_prior}) averaged over three separate runs and a set of $\{1, 12\}$ MC samples.
For the \textbf{beta scale} ablations, we favored the $\mathcal{N}(0, 1)$ based on its superior performance. The results, once again, were averaged over three distinct runs using the same MC sample set, ${1, 12}$.
Lastly, for the number of \textbf{Monte Carlo samples} ablation, we employed the $\mathcal{N}(0, 1)$ prior, showcasing results for both 1 and 12 MC samples, with their averages derived from three unique runs.

\section{Probabilistic embeddings}
\label{app:probabilistic_embeddings}

\subsection{Variance of the embeddings}
\label{app:embeddings_variance}

\begin{figure}[t]
    \centering
    \includegraphics[width=0.49\textwidth]{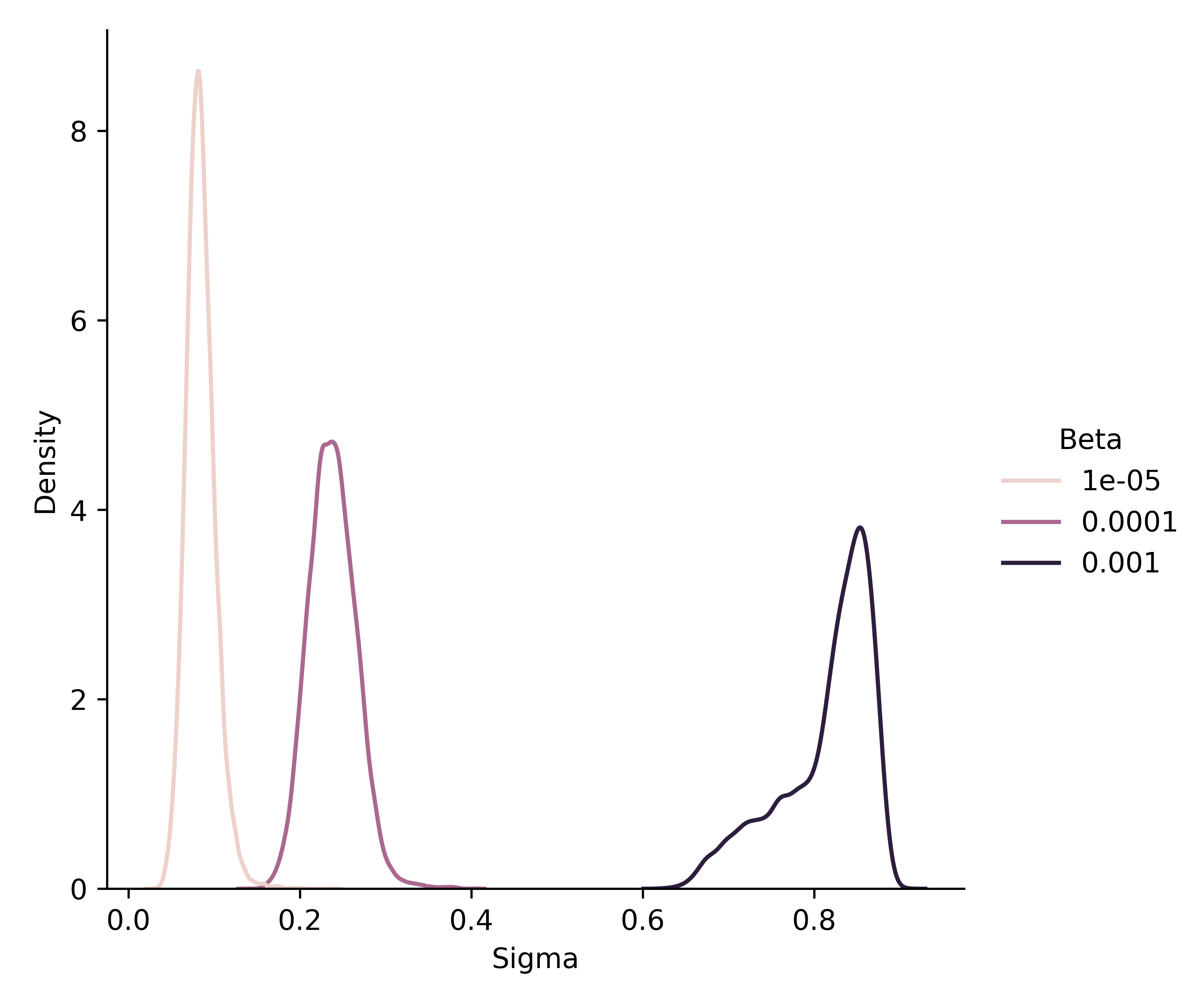}
    \caption{Probabilistic embeddings' variance for $\text{H-prob.}$ VICReg model.}
    \label{fig:ssl_bayes_h_variance}
\end{figure}

\begin{figure}[h]
    \centering
    \includegraphics[width=0.49\textwidth]{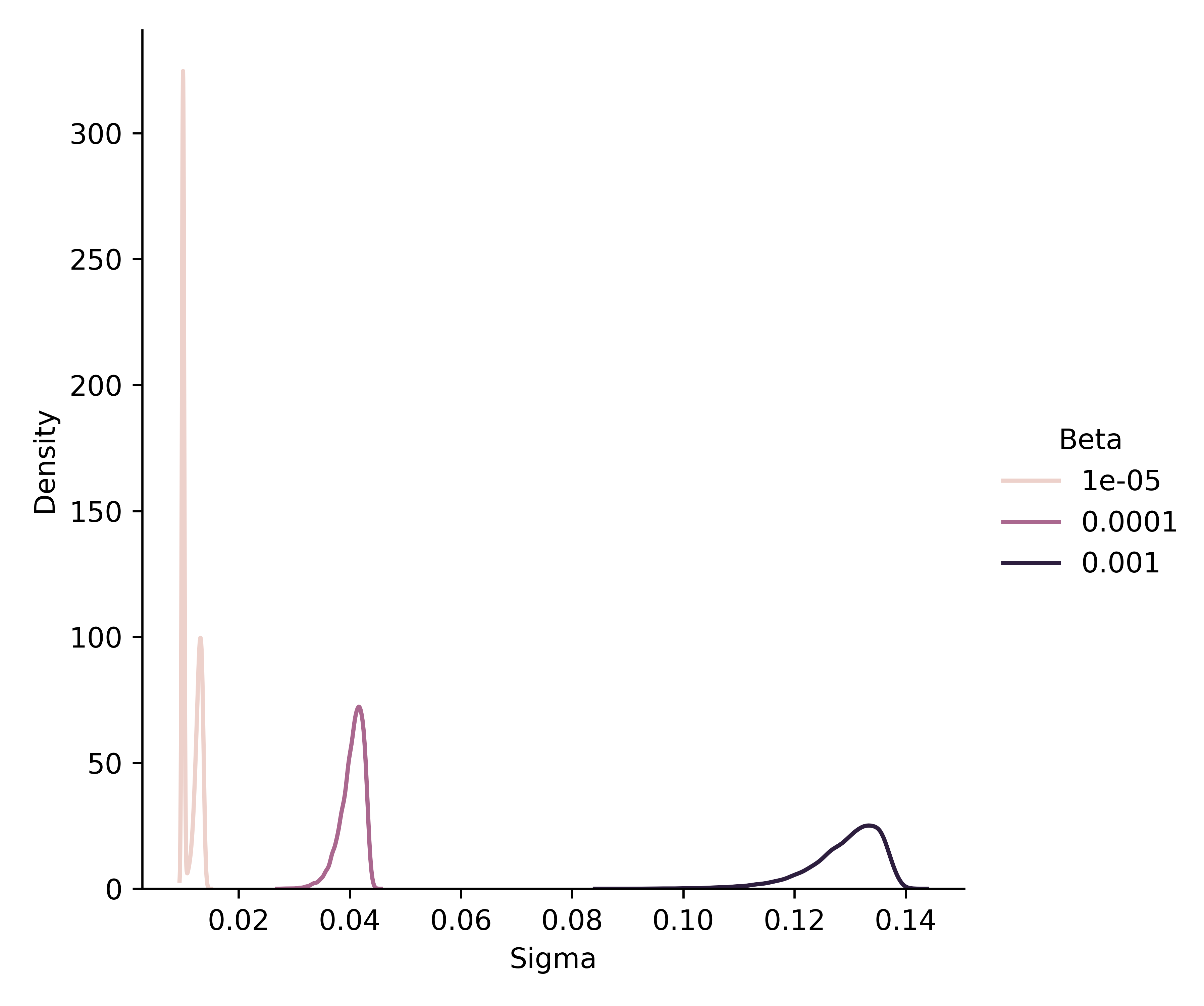}
    \caption{Probabilistic embeddings' variance for $\text{Z-prob.}$ VICReg model.}
    \label{fig:ssl_bayes_z_variance}
\end{figure}

In Section \ref{sec:ablations}, we highlighted that decreasing the value of beta correspondingly reduces the variance of the embeddings, pushing them towards deterministic embeddings. 
To illustrate this, we plotted the estimated density of the mean variance of the embeddings in Figures \ref{fig:ssl_bayes_h_variance} and \ref{fig:ssl_bayes_z_variance}. These plots are based on models from the ablation study, which were trained on the CIFAR-10 dataset.
We have observed that by setting a particularly low value for $\beta$, the embeddings tend to be almost entirely deterministic. While this is not ideal — as it negates the benefits of stochasticity inherent to probabilistic embeddings — there exists a balance. The challenge lies in minimizing the variance of the embeddings, which typically enhances performance while maintaining sufficient stochasticity to harness the benefits of their probabilistic nature.

\begin{figure}[h]
    \centering
    \includegraphics[width=0.49\textwidth]{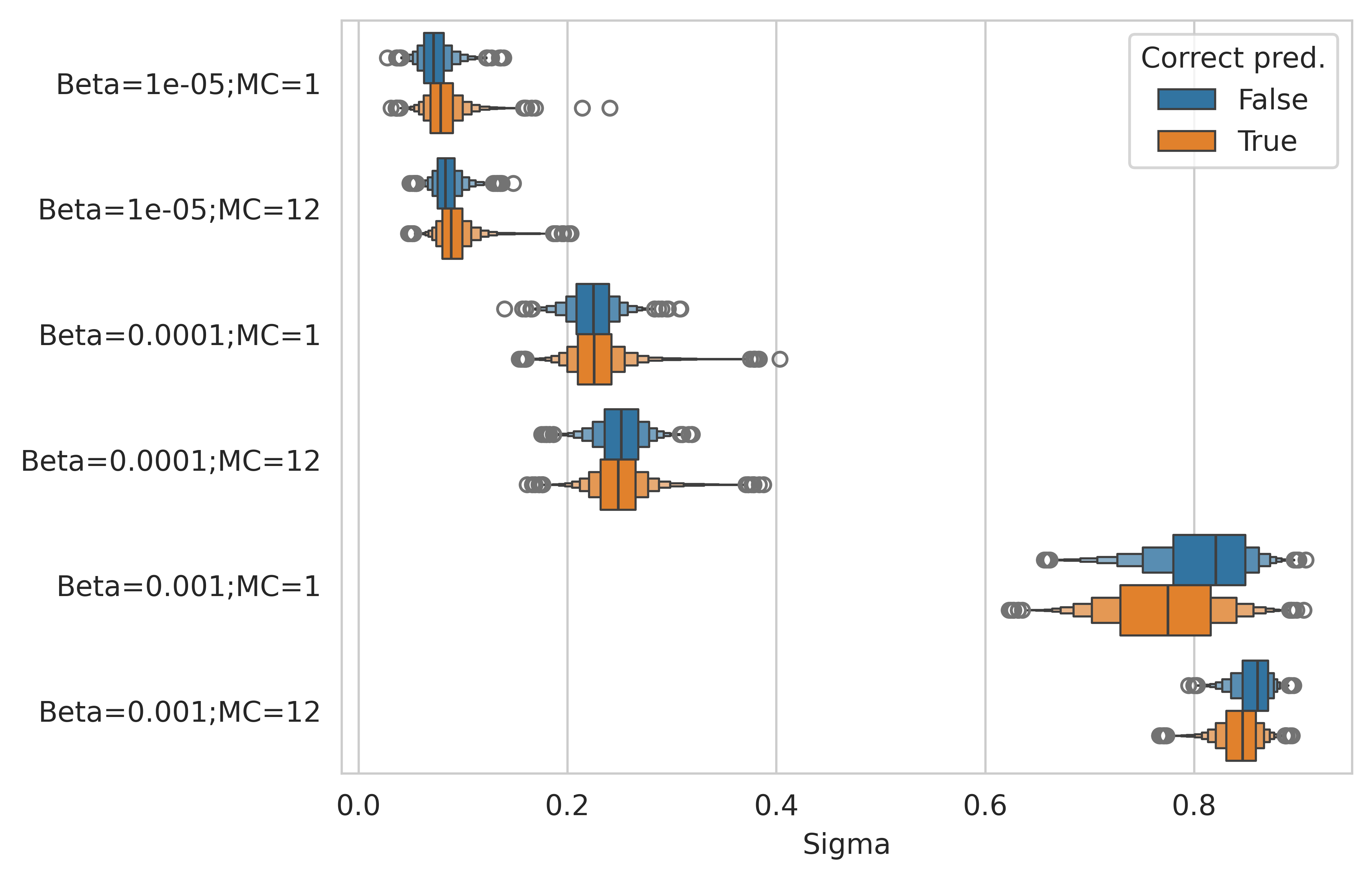}
    \caption{Mean sigma for correct and incorrect examples for $\text{H-prob.}$ embeddings and VICReg method.}
    \label{fig:simga_vs_pred_ssl_bayes_h}
\end{figure}

\subsection{IN-distribution uncertainties}
\begin{figure}[h]
    \centering
    \includegraphics[width=0.49\textwidth]{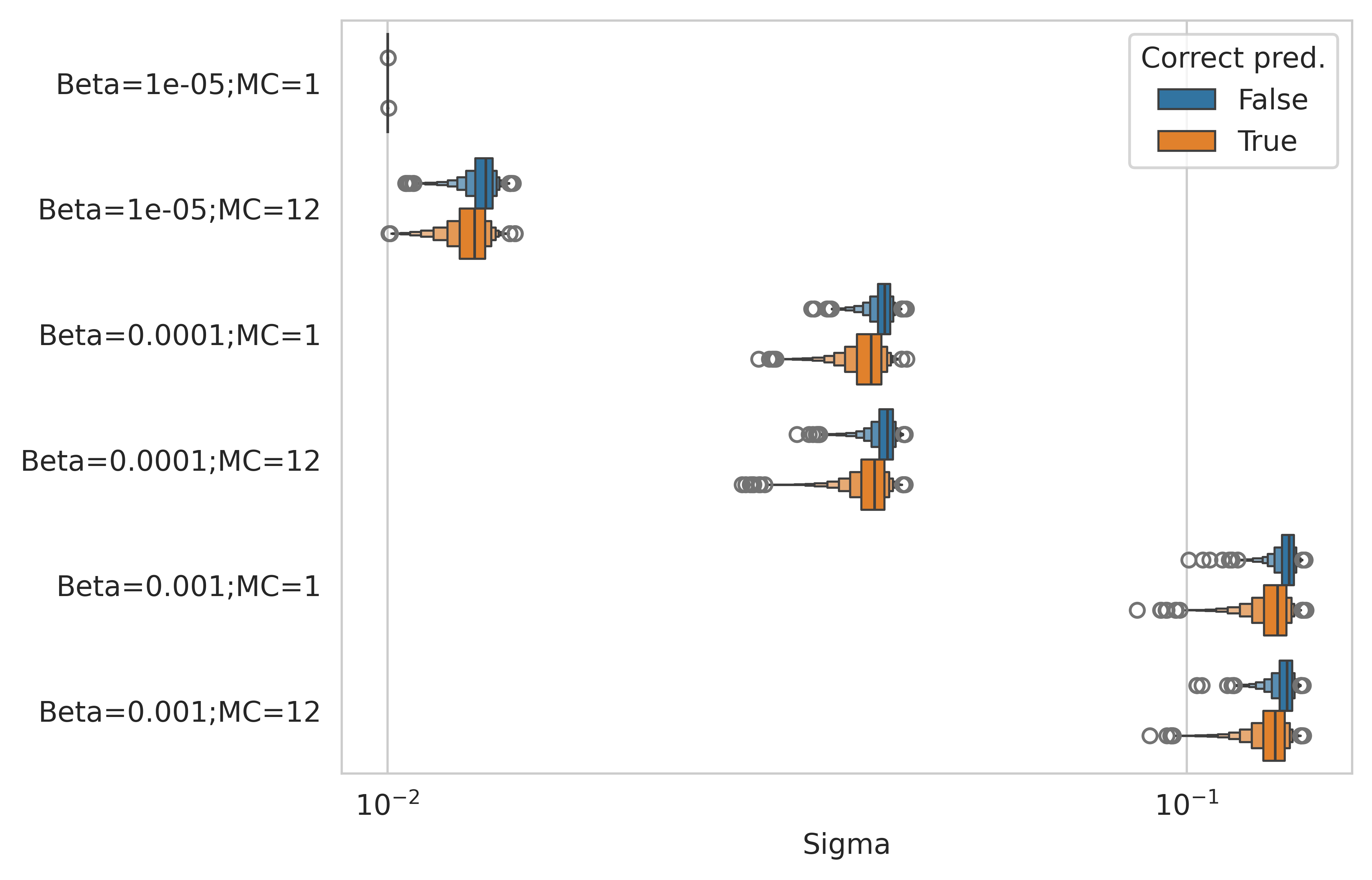}
    \caption{Mean sigma variance for correct and incorrect examples for  $\text{Z-prob.}$ embeddings and VICReg method.}
    \label{fig:simga_vs_pred_ssl_bayes_z}
\end{figure}

We delved into the relationship between uncertainty estimates and model predictions, aiming to discern if the variance of the embeddings (sigma) could serve as an uncertainty measure for examples from the in-distribution dataset. Our specific interest was to see if the variance would be higher for more challenging examples where the model is prone to making errors. Figures \ref{fig:simga_vs_pred_ssl_bayes_h} and \ref{fig:simga_vs_pred_ssl_bayes_z} illustrate the mean variance values across different model variants on the CIFAR-10 dataset.
We can observe that, in most cases, the distribution of mean variance of the embeddings is actually shifted, with incorrect predictions often being assigned a higher variance.

\section{Out-of-distribution detection}
\label{app:ood}

\begin{table*}[t]
\caption{Results for various hyperparameter settings in the OOD detection task using VICReg.}
\label{tab:app_ood_main_vicreg}
\begin{center}
\begin{adjustbox}{width=1\textwidth}
\begin{tabular}{llrlrrrrrrlrrrrrr}
\hline
\textbf{}   &  & Deterministic & \textbf{} & \multicolumn{6}{c}{Z-prob.}                            & \textbf{} & \multicolumn{6}{c}{Prob-H}                             \\ \cline{1-3} \cline{5-10} \cline{12-17} 
\textbf{}   &  & N/A           & \textbf{} & \multicolumn{3}{c}{Standard} & \multicolumn{3}{c}{MoG} & \textbf{} & \multicolumn{3}{c}{Standard} & \multicolumn{3}{c}{MoG} \\ \cline{1-3} \cline{5-10} \cline{12-17} 
\textbf{}   &  & N/A           & \textbf{} & 1e-5     & 1e-4    & 1e-3    & 1e-5   & 1e-4   & 1e-3  & \textbf{} & 1e-5     & 1e-4    & 1e-3    & 1e-5   & 1e-4   & 1e-3  \\ \hline \hline
Entropy     &  & 0.827         &           & 0.813    & 0.827   & 0.813   & 0.826  & 0.819  & 0.756 &           & 0.817    & 0.804   & 0.800   & 0.826  & 0.821  & 0.819 \\
MaxSoftmax  &  & 0.799         &           & 0.784    & 0.800   & 0.789   & 0.801  & 0.794  & 0.723 &           & 0.790    & 0.783   & 0.781   & 0.798  & 0.794  & 0.792 \\
ODIN        &  & 0.788         &           & 0.772    & 0.793   & 0.780   & 0.793  & 0.786  & 0.740 &           & 0.776    & 0.765   & 0.770   & 0.788  & 0.793  & 0.776 \\ \hline
Mahalanobis &  & 0.649         &           & 0.655    & 0.662   & 0.689   & 0.664  & 0.670  & 0.673 &           & 0.671    & 0.732   & 0.646   & 0.646  & 0.735  & 0.583 \\
SigmaMean   &  & N/A           &           & 0.671    & 0.753   & 0.753   & 0.664  & 0.760  & 0.628 &           & 0.583    & 0.654   & 0.701   & 0.640  & 0.768  & 0.695 \\
SigmaStd    &  & N/A           &           & 0.777    & 0.781   & 0.747   & 0.776  & 0.633  & 0.609 &           & 0.635    & 0.655   & 0.648   & 0.643  & 0.604  & 0.707 \\ \hline
\end{tabular}
\end{adjustbox}
\end{center}
\end{table*}

\begin{table*}[h]
\caption{Results for various hyperparameter settings in the OOD detection task using Barlow Twins.}
\label{tab:app_ood_main_barlow}
\begin{center}
\begin{adjustbox}{width=1\textwidth}
\begin{tabular}{lrlrrrrrrlrrrrrr}
\hline
\textbf{}   & Deterministic & \textbf{} & \multicolumn{6}{c}{Z-prob.}                            & \textbf{} & \multicolumn{6}{c}{H-prob.}                            \\ \cline{1-2} \cline{4-9} \cline{11-16} 
\textbf{}   & N/A           & \textbf{} & \multicolumn{3}{c}{Standard} & \multicolumn{3}{c}{MoG} & \textbf{} & \multicolumn{3}{c}{Standard} & \multicolumn{3}{c}{MoG} \\ \cline{1-2} \cline{4-9} \cline{11-16} 
            & N/A           & \textbf{} & 1e-03    & 1e-02   & 1e-01   & 1e-03  & 1e-02  & 1e-01 & \textbf{} & 1e-04    & 1e-03   & 1e-02   & 1e-04  & 1e-03  & 1e-02 \\ \hline \hline
Entropy     & 0.823         &           & 0.818    & 0.828   & 0.814   & 0.827  & 0.818  & 0.771 &           & 0.820    & 0.817   & 0.808   & 0.819  & 0.828  & 0.828 \\
MaxSoftmax  & 0.795         &           & 0.791    & 0.801   & 0.789   & 0.801  & 0.792  & 0.742 &           & 0.795    & 0.789   & 0.787   & 0.792  & 0.797  & 0.797 \\
ODIN        & 0.783         &           & 0.782    & 0.793   & 0.782   & 0.790  & 0.784  & 0.748 &           & 0.793    & 0.783   & 0.785   & 0.786  & 0.794  & 0.794 \\ \hline
Mahalanobis & 0.654         &           & 0.697    & 0.655   & 0.694   & 0.698  & 0.692  & 0.719 &           & 0.678    & 0.638   & 0.644   & 0.646  & 0.627  & 0.631 \\
SigmaMean   &    N/A           &           & 0.631    & 0.639   & 0.628   & 0.650  & 0.642  & 0.614 &           & 0.621    & 0.580   & 0.640   & 0.620  & 0.586  & 0.634 \\
SigmaStd    &       N/A        &           & 0.800    & 0.782   & 0.717   & 0.756  & 0.805  & 0.620 &           & 0.623    & 0.690   & 0.647   & 0.646  & 0.673  & 0.594 \\ \hline
\end{tabular}
\end{adjustbox}
\end{center}
\end{table*}

In this section, we present a comprehensive set of results for the out-of-distribution detection task. The evaluation methodology remains consistent with that described in Section \ref{sec:ood}. Tables \ref{tab:app_ood_main_vicreg} and \ref{tab:app_ood_main_barlow} display the results for both the VICReg and Barlow Twins methods, taking into account various hyperparameters, including the choice of prior, the beta ($\beta$) scale, and the number of MC samples. The results have been averaged over three runs using distinct seeds and across all OOD datasets specified in Section \ref{sec:ood} (Textures \cite{cimpoi_describing_2013}, TinyImageNet(crop, resized) \cite{le_tiny_2015} and LSUN(crop, resized) \cite{yu_lsun_2016}). 

The AUROC performance for OOD detection offers insightful comparisons between various methods. Specifically, Entropy \cite{chan_entropy_2021}, MaxSoftmax \cite{hendrycks_baseline_2018}, and ODIN \cite{liang_enhancing_2020} are techniques that require label information to detect out-of-distribution examples. On the other hand, methods such as Mahalanobis \cite{lee_simple_2018}, SigmaMean, and SigmaStd are based solely on representations and do not necessitate label information for OOD detection.
This distinction underscores the variety of approaches available in the field, ranging from those dependent on labeled data to others that leverage unsupervised information.

For the VICReg method, an intermediate value of $\beta$ frequently yields the best performance across detectors. With optimal values of beta (i.e., $1e-4$), the representation gains from the added bottleneck created by stochastic embeddings. Moreover, we can leverage the characteristics of stochastic embeddings, specifically its variance, as an OOD predictor. 
Interestingly, the Mahalanobis detector benefits greatly from $\text{H-prob.}$ embeddings for such beta for both Standard and MoG priors. 
Meanwhile, Table \ref{tab:app_ood_main_barlow} presents the results specific to the Barlow Twins method. Notably, the SigmaStd detector consistently outperforms the SigmaMean detector in nearly all scenarios, and it is particularly effective with the $\text{Z-prob.}$ embeddings.
The performance of $\text{Z-prob.}$ embeddings in the out-of-distribution detection task are particularly impressive, matching or surpassing detectors relying on label information in both the Barlow Twins and VICReg methods.

\begin{table*}[t]
\caption{Mutual information, loss values and accuracy for different hyperparameter settings using Barlow Twins.}
\label{tab:mutual_information}
\begin{center}
\begin{adjustbox}{width=1\textwidth}
\begin{tabular}{lllrrrrrrr}
\toprule
Embeddings & Beta ($\beta$) & MC samples &  Acc@1 &  Reg. loss &  Inv. loss &  $I(V;H)$ &  $I(H;H')$ &  $I(H;Z)$ &  $I(Z;Z')$ \\
\midrule \midrule
      Deterministic &   N/A         &      N/A      &  0.833 &       69.7 &       50.3 &    3.37 &     4.02 &    3.74 &     3.74 \\ \hline
   H-prob. &     0.0001 &          1 &  0.801 &       78.4 &       54.4 &    3.31 &     3.87 &    7.25 &     3.68 \\
   H-prob. &     0.0001 &         12 &  0.803 &        108 &       53.2 &    3.37 &     4.22 &    9.89 &     3.99 \\
   H-prob. &      0.001 &          1 &  0.799 &       79.3 &       55.5 &    3.34 &     3.97 &    6.66 &     3.62 \\
   H-prob. &      0.001 &         12 &  0.809 &        112 &       51.6 &    3.32 &     4.31 &    9.54 &     3.99 \\
   H-prob. &       0.01 &          1 &   0.79 &       77.8 &       56.6 &    3.08 &     3.82 &    6.34 &     3.42 \\
   H-prob. &       0.01 &         12 &  0.805 &       95.8 &       52.9 &    3.08 &     4.17 &    8.94 &     3.94 \\ \hline
   Z-prob. &      0.001 &          1 &  0.823 &       73.9 &       50.5 &    3.52 &     4.06 &    6.68 &     3.86 \\
   Z-prob. &      0.001 &         12 &  0.826 &       76.6 &       52.1 &    3.56 &     3.94 &    7.08 &     4.02 \\
   Z-prob. &       0.01 &          1 &  0.827 &       73.4 &       55.1 &     3.5 &     4.05 &     6.5 &     3.86 \\
   Z-prob. &       0.01 &         12 &  0.827 &       75.1 &       56.7 &     3.5 &     3.93 &    7.45 &     4.03 \\
   Z-prob. &        0.1 &          1 &  0.821 &       78.9 &       93.2 &    3.39 &        4 &    6.34 &     3.68 \\
   Z-prob. &        0.1 &         12 &  0.819 &         83 &         96 &    3.45 &     3.82 &    6.91 &     3.87 \\
\bottomrule
\end{tabular}
\end{adjustbox}
\end{center}
\end{table*}

\section{Information bottleneck}
\label{app:information_bottleneck}

We employ the Mutual Information Neural Estimation (MINE) technique as detailed in \cite{belghazi_mine_2021} to evaluate the mutual information between the input, representation, and embeddings. For every pair of variables, a distinct network is designated to quantify their mutual information. This network is trained jointly alongside the primary self-supervised network, utilizing a separate optimizer. The \textit{statistic network}, as referred to in MINE, in our implementation, comprises two layers, each with a dimensionality of 1024. These layers are succeeded by a ReLU nonlinearity and then followed by a third layer that maps to a singular output. The values obtained, showcased in Table \ref{tab:mutual_information}, correspond to the results from the final epoch.

From our observations, the mutual information $I(V;H)$ between the input and representation varies across different embeddings. A larger beta value results in a decreased $I(V;H)$ for the $\text{H-prob.}$ embeddings, yet it increases for every variant of the $\text{Z-prob.}$ embeddings. Increasing the number of MC samples improves both $I(H;H)$ and $I(Z;Z)$. These can be interpreted as a lower bound to the \textit{predictive information} on an unknown label $y$, as cited in \cite{federici_learning_2020}. Furthermore, the mutual information $I(H;Z)$ between representation and the loss space is considerably elevated for probabilistic embeddings. A detailed discussion on this phenomenon is provided in Section \ref{sec:info_compression}.

\end{document}